\documentclass{article}

\usepackage{mathtools}
\usepackage{multirow}
\usepackage{xcolor,graphicx}
\usepackage{csquotes}

\usepackage[font=small,labelfont=bf]{caption}
\DeclareCaptionLabelSeparator{separator}{\textbf{.}\enspace}
\usepackage{hyperref}
\hypersetup{colorlinks,citecolor=black,filecolor=black,linkcolor=black,urlcolor=black}

\usepackage{fancyhdr,lipsum,setspace,floatpag}
\floatpagestyle{fancy}

\usepackage[style=nature,alldates=year,backend=bibtex,doi=false,isbn=false,url=false]{biblatex}
\bibliography{HNstudy_SciRep2017}

\begin{document}
\thispagestyle{plain}

\begin{titlepage}

\title{\textbf{\Large{Radiomics strategies for risk assessment of tumour failure in head-and-neck cancer}}}
\date{}
\maketitle

\noindent \footnotesize{\author{\textbf{Martin Valli\`eres}$^{1,*}$, \textbf{Emily Kay-Rivest}$^2$, \textbf{L\'eo Jean Perrin}$^3$, \textbf{Xavier Liem}$^4$, \textbf{Christophe Furstoss}$^5$, \textbf{Hugo J. W. L. Aerts}$^6$, \textbf{Nader Khaouam}$^5$, \textbf{Phuc Felix Nguyen-Tan}$^4$, \textbf{Chang-Shu Wang}$^3$, \textbf{Khalil Sultanem}$^2$, \textbf{Jan Seuntjens}$^1$ \& \textbf{Issam El Naqa}$^7$}} \\ \\
\scriptsize{
$^1$ Medical Physics Unit, McGill University, Cedars Cancer Centre, McGill University Health Centre -- Glen Site, 1001 boulevard D\'ecarie, Montr\'eal, QC H4A 3J1, Canada. \\
$^2$ Radiation Oncology Division, H\^opital g\'en\'eral juif, 3755 C\^ote Ste. Catherine, Montr\'eal, QC H3T 1E2, Canada \\
$^3$ Department of Radiation Oncology, Centre hospitalier universitaire de Sherbrooke, H\^opital Fleurimont, 3001 12$^{\mathrm{e}}$ avenue N, Sherbrooke, QC J1H 5N4, Canada  \\
$^4$ Department of Radiation Oncology, Centre hospitalier de l'Universit\'e de Montr\'eal, H\^opital Notre-Dame, 1560 Sherbrooke E, Montr\'eal, QC H2L 4M1, Canada \\
$^5$ Department of Radiation Oncology, H\^opital Maisonneuve-Rosemont, 5415 boulevard Assomption, Montr\'eal, QC H1T 2M4, Canada \\
$^6$ Departments of Radiation Oncology \& Radiology, Dana-Farber Cancer Institute, Brigham and Women's Hospital, Harvard Medical School, Boston, Massachusetts 02215-5450, USA \\
$^7$ Department of Radiation Oncology, Physics Division, University of Michigan, 519 W. Williman St, Argus Bldg, Ann Arbor, Michigan 48103-4943, USA} \\

\noindent *Corresponding author: mart.vallieres@gmail.com
\bigskip

\begin{abstract}
Quantitative extraction of high-dimensional mineable data from medical images is a process known as radiomics. Radiomics is foreseen as an essential prognostic tool for cancer risk assessment and the quantification of intratumoural heterogeneity. In this work, 1615 radiomic features (quantifying tumour image intensity, shape, texture) extracted from pre-treatment FDG-PET and CT images of 300 patients from four different cohorts were analyzed for the risk assessment of locoregional recurrences (LR) and distant metastases (DM) in head-and-neck cancer. Prediction models combining radiomic and clinical variables were constructed via random forests and imbalance-adjustment strategies using two of the four cohorts. Independent validation of the prediction and prognostic performance of the models was carried out on the other two cohorts (LR: $\mathrm{AUC} = 0.69$ and $\mathrm{CI} = 0.67$; DM: $\mathrm{AUC} = 0.86$ and $\mathrm{CI} = 0.88$). Furthermore, the results obtained via Kaplan-Meier analysis demonstrated the potential of radiomics for assessing the risk of specific tumour outcomes using multiple stratification groups. This could have important clinical impact, notably by allowing for a better personalization of chemo-radiation treatments for head-and-neck cancer patients from different risk groups.
\end{abstract}
\bigskip 

\noindent \textit{Preprint submitted to \enquote{Scientific Reports} on March 14, 2017}

\thispagestyle{empty}
\end{titlepage}
\setcounter{page}{2}

\onehalfspacing 

\section*{Introduction}

\noindent Precision oncology promises to tailor the full spectrum of cancer care to an individual patient, notably in terms of personalization of cancer prevention, screening, risk stratification, therapy and response assessment. With sufficient infrastructure support and concerted efforts from the different stakeholders, it is possible to foresee that personalized therapy would become the standard of care in oncology \autocite{meric-bernstam_building_2013}. Cancer mechanisms are increasingly elucidated as functions of different biomarkers or tumour genetic mutations, thereby changing the way we design clinical trials to achieve better cancer management efficacy in specific patient sub-populations \autocite{renfro_precision_2016}. On the other hand, \enquote{rapid learning paradigms} (i.e. knowledge-driven healthcare) consisting of reusing routine clinical data to develop knowledge in the form of models that can predict treatment outcomes for a larger portion of the population have also gained popularity in the oncology community \autocite{lambin_rapid_2013,shrager_rapid_2014}. Although most research approaches to precision oncology are centered on genomics technologies \autocite{weitzel_genetics_2011,garraway_precision_2013}, it is thought that only the integration of multiple-omics, i.e., panomics data (genomics, transcriptomics, proteomics, metabolomics, etc.) could efficiently unravel biological mechanisms \autocite{el_naqa_biomedical_2014,ebrahim_multi-omic_2016}.

The importance of panomics integration for cancer risk assessment emerges from the tremendous extent of heterogeneous characteristics expressed at multiple levels of tumours. Genes, proteins, cellular microenvironments, tissues and anatomical landmarks within tumours exhibit considerable spatial and temporal variations that could potentially yield valuable information about tumour aggressiveness. Tumours are generally composed of multiple clonal sub-populations of cancer cells forming complex dynamic systems that exhibit rapid evolution as a result of their interaction with their microenvironment and therapy perturbations \autocite{fisher_cancer_2013}. Differing properties can be attributed to the different sub-populations in terms of growth rate, expression of biomarkers, ability to metastasize, and immunological characteristics \autocite{heppner_tumor_1983}. These properties could be described by differences in metabolic activity, cell proliferation, oxygenation levels, pH, blood vasculature and necrotic areas observed within the tumour. Such intratumoural differences are related to the concept of tumour heterogeneity, a characteristic that can be observed with significantly different extents even amongst tumours of the same histopathological type. Tumours exhibiting such heterogeneous characteristics are thought to be associated with high risk of resistance to treatment, progression, metastasis or recurrence \autocite{fidler_critical_1990,yokota_tumor_2000,campbell_patterns_2010}.

Nowadays, medical imaging plays a central role in the investigation of intratumoural heterogeneity, as radiological images are acquired as routine practice for almost every patient with cancer. Medical images such as 2-deoxy-2-[$^{18}$F]fluoro-D-glucose (FDG) positron emission tomography (PET) and X-ray computed tomograph (CT) are minimally invasive and they carry an immense source of potential data for decoding the tumour phenotype \autocite{gillies_radiomics:_2016}. The quantitative extraction of high-dimensional mineable data from all types of medical images and whose subsequent analysis aims at supporting clinical decision-making is a process coined with the term \enquote{radiomics} \autocite{el_naqa_exploring_2009,gillies_biology_2010,lambin_radiomics:_2012,kumar_radiomics:_2012}. The demonstration that gene-expression signatures and clinical phenotypes could be inferred from tumour imaging features \autocite{segal_decoding_2007,diehn_identification_2008,aerts_decoding_2014} has led to an exponential growth of this field in the past few years \autocite{hatt_characterization_2016,yip_applications_2016}. The underlying hypothesis of radiomics is that the genomic heterogeneity of aggressive tumours could translate into heterogeneous tumour metabolism and anatomy, thereby envisioning the quantitative analysis of diagnostic medical images as an essential prognostic tool for cancer risk assessment and as an integral part of panomic tumour signature profiling.

\begin{figure}[!htbp]
	\includegraphics[width=\textwidth]{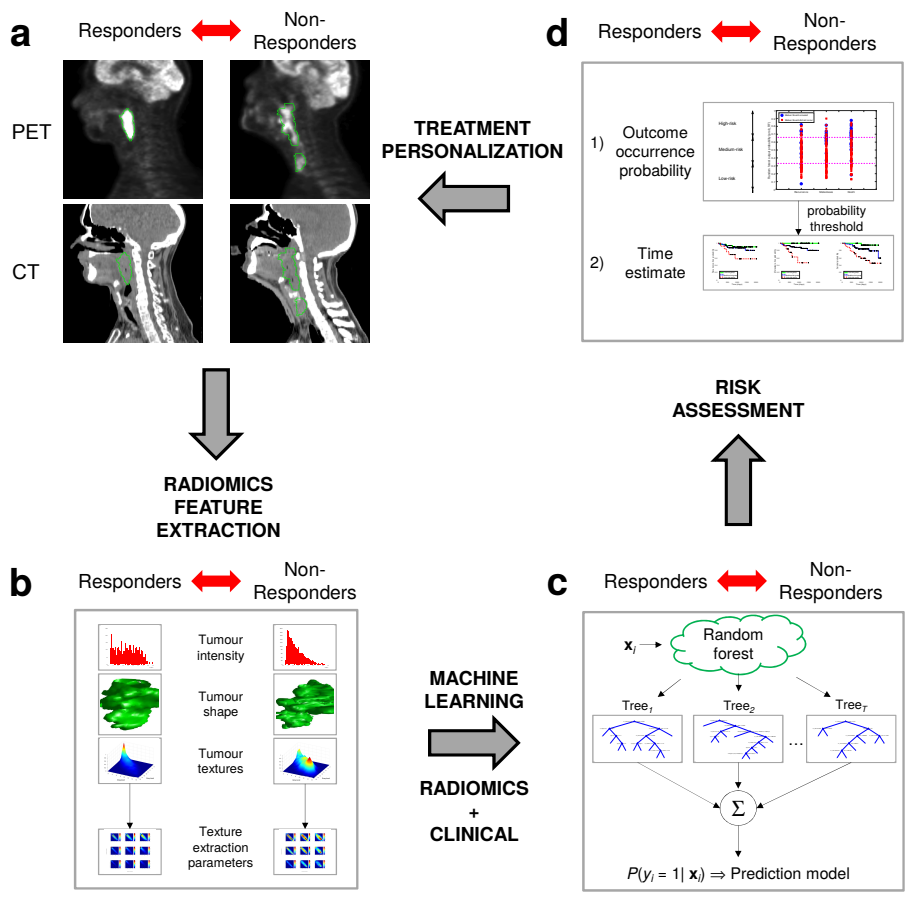}
	\caption{\textbf{From radiomics analysis to treatment personalization.} \footnotesize{\textbf{(a)} Example of diagnostic FDG-PET and CT images of two head-and-neck cancer patients with tumour contours. The patient that did not respond well to treatment (right) has a more heterogeneous intratumoural intensity distribution in both FDG-PET and CT images than the patient that responded well to treatment (left). \textbf{(b)} The radiomics analysis strategy involves the extraction of features differentiating responders from non-responders to treatment. Features are extracted from the FDG-PET and CT tumour contours and quantify tumour shape, intensity, and texture. \textbf{(c)} Advanced machine learning combines radiomics features and patient clinical information via a random forest algorithm. The classifier is trained to differentiate between responders and non-responders to treatment (prediction model). \textbf{(d)} The output probability of the random forest classifier computed on new patients can be used to assess the risk of non-response to treatment via probabilities of occurrence of outcome events and time estimates. Eventually, accurate risk assessment of specific tumour outcomes via radiomics analysis could help to better personalize cancer treatments.}} \label{fig:StudyWorkflow}
\end{figure}

The translation of radiomics analysis into standard cancer care to support treatment decision-making involves the development of prediction models integrating clinical information that can assess the risk of specific tumour outcomes \autocite{lambin_predicting_2013}(Fig.~\ref{fig:StudyWorkflow}). In this work, our main objective is to construct prediction models using advanced machine learning to evaluate the risk of locoregional recurrences and distant metastases prior to chemo-radiation of head-and-neck (H\&N) cancers, a group of biologically similar neoplasms originating from the squamous cells that line the mucosal surfaces in the oral cavity, paranasal sinuses, pharynx or larynx. The locoregional control of H\&N cancers is usually good, but this is, however, not matched by improvements in survival, as the development of distant metastases and second primary cancers are the leading causes of treatment failure and death \autocite{ferlito_incidence_2001,baxi_causes_2014}. In order to improve patient survival and outcomes, the importance of identifying relevant prognostic factors that can better assess the aggressiveness of tumours at the moment of diagnosis is crucial. 

\begin{figure}[!htbp]
	\centering
	\includegraphics[width=\textwidth]{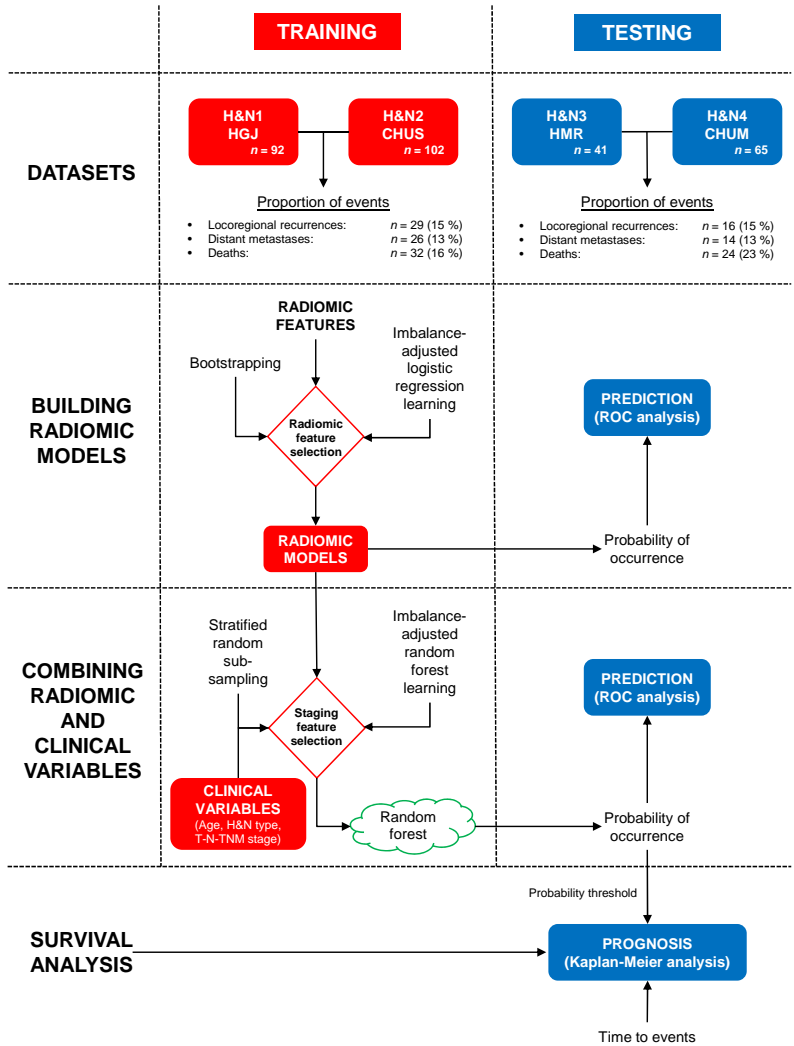}
	\caption{\textbf{Models construction strategy and analysis workflow.} \footnotesize{Four different cohorts were used to demonstrate the utility of radiomics analysis for the pre-treatment assessment of the risk of locoregional recurrence and distant metastases in head-and-neck cancer. The H$\&$N1 and H$\&$N2 cohorts were combined and used as a single training set ($n=194$), whereas the H$\&$N3 and H$\&$N4 cohorts were combined and used as a single testing set ($n=106$). The best combinations of radiomics features were selected in the training set using imbalance-adjusted logistic regression learning and bootstrapping validations. These radiomics features were combined with selected clinical variables in the training set using imbalance-adjusted random forest learning and stratified random sub-sampling validations. Independent prediction analysis was performed in the testing set for all classifiers fully constructed in the training set. Independent prognosis analysis and Kaplan-Meier risk stratification was carried out in the testing set using the output probability of occurrence of events of random forests fully constructed in the training set.}} \label{fig:MLstrategy}
\end{figure}

We hypothesize that radiomic features are important prognostic factors for the risk assessment of specific H\&N cancer outcomes \autocite{wong_radiomics_2016}. The machine learning strategy employed in this work involves the extraction of 1615 different radiomic features from a total of 300 patients from four different institutions. Two cohorts are used to construct the prediction models by combining radiomics (intensity, shape, textures) and clinical attributes (patient age, H\&N type, tumour stage) via random forests classifiers and imbalance-adjustments of training samples, and the remaining two cohorts are reserved to evaluate the prediction (binary assessment of outcome) and prognostic (time-to-event assessment) performance of the corresponding models (Fig.~\ref{fig:MLstrategy}). Throughout this study, results obtained for locoregional recurrences and distant metastases are also compared against prediction models constructed for the general risk assessment of overall survival in H\&N cancer. A comprehensive comparison of the prediction/prognostic performance of radiomics versus clinical models and volumetric variables is also performed. Our results suggest that the integration of radiomic features into clinical prediction models has considerable potential for assessing the risk of specific outcomes prior to treatment of H\&N cancers. Accurate stratification of locoregional recurrence and distant metastasis risks could eventually provide a rationale for adapting the radiation doses and chemotherapy regimens that the patients receive. Overall, combining quantitative imaging information with other categories of prognostic factors via advanced machine learning could have a profound impact on the characterization of tumour phenotypes and would increase the possibility of translation of outcome prediction models into the clinical environment as a means to personalize treatments.

\bigskip

\newpage

\section*{Results}

\subsection*{Summary of presentation of results}
To ease reading and the understanding of this study, a summary of how results are presented in the text is provided in Supplementary Fig.~S1.

\subsection*{Association of variables with tumour outcomes}
In order to assess the value of quantitative pre-treatment imaging to predict specific cancer outcomes in H\&N cancer, we performed a comprehensive univariate analysis of the association of radiomic features with locoregional recurrences (\enquote{LR} or \enquote{Locoregional}), distant metastases development (\enquote{DM} or \enquote{Distant}) and overall survival (\enquote{OS} or \enquote{Survival} or \enquote{Death}). A total of 1615 radiomic features (Fig.~\ref{fig:StudyWorkflow}b and Supplementary Methods section 2.5 for complete description) were first extracted from the gross tumour volume (GTV$_{\mathrm{primary}}$ $+$ GTV$_{\mathrm{lymph\ nodes}}$) of the FDG-PET and CT images (Fig.~\ref{fig:StudyWorkflow}a), for all 300 patients from the four H\&N cancer cohorts (Fig.~\ref{fig:MLstrategy}): I) 10 first-order statistics features (intensity); II) 5 morphological features (shape); and III) 40 texture features each extracted using 40 different combinations of parameters. We also compared these results to the predictive power of the tumour volume (\enquote{\textit{Volume}}) and of the following clinical variables: \textit{Age}, \textit{T-Stage}, \textit{N-Stage}, \textit{TNM-Stage} and human papillomavirus status (\textit{HPV status}), where \textit{HPV status} was available for 120 of the 300 patients (Supplementary Tables~S5-S8). The association of the different variables with the different H\&N cancer outcomes (binary endpoints) was then analyzed using Spearman's rank correlations ($r_s$) computed on all patients, and significance was assessed by applying multiple testing corrections using the Benjamini-Hochberg procedure \autocite{benjamini_controlling_1995} with a false discovery rate of 10~\%.

Overall, we found that 0~\%, 63~\% and 12~\% of the total radiomic features extracted from PET scans, and that 0~\%, 61~\% and 34~\% of the total radiomic features extracted from CT scans were significantly associated with LR, DM and OS, respectively (after multiple testing corrections). The radiomic features (PET or CT) with the highest associations with LR, DM and OS were $LZHGE_{\mathrm{GLSZM}}$ from CT scans ($r_s = -0.15$, $p = 0.007$), $ZSN_{\mathrm{GLSZM}}$ from CT scans ($r_s = -0.29$, $p = 2 \times 10^{-7}$) and $GLV_{\mathrm{GLRLM}}$ from CT scans ($r_s = 0.24$, $p = 4 \times 10^{-5}$), respectively (Supplementary Table~S1). Tumour volume was not found to be significantly associated with LR ($r_s = -0.04$, $p = 0.48$), but was significantly associated with DM ($r_s = 0.24$, $p = 3 \times 10^{-5}$) and OS ($r_s = -0.18$, $p = 2 \times 10^{-3}$). Finally, we found that \{\textit{Age}, \textit{T-Stage}, \textit{N-Stage}, \textit{HPV status}\}, \{\textit{N-Stage}\} and \{\textit{Age}, \textit{T-Stage}, \textit{HPV status}\} were significantly associated with LR, DM and OS, respectively. The clinical variables with the highest associations with LR, DM and OS were \textit{HPV}$-$ ($r_s = 0.39$, $p = 8 \times 10^{-6}$), higher \textit{N-Stage} ($r_s = 0.18$, $p = 1 \times 10^{-3}$) and higher \textit{T-Stage} ($r_s = 0.21$, $p = 3 \times 10^{-4}$), respectively (Supplementary Table~S2).

\subsection*{Construction of prediction models}
The construction of prediction models for LR, DM and OS was carried out using a training set consisting of the combination of 194 patients from the H\&N1 and H\&N2 cohorts (Fig.~\ref{fig:MLstrategy}). Three initial radiomic feature sets were considered: I) the 1615 radiomic features extracted from PET scans (\enquote{\textit{PET}} feature set); II) the 1615 radiomic features extracted from CT scans (\enquote{\textit{CT}} feature set); and III) a combined set containing all PET and CT radiomic features used in feature sets I and II (\enquote{\textit{PETCT}} feature set).

Prediction models consisting of radiomic information only were first constructed for each of the three H\&N outcomes and the three initial radiomic feature sets. \textit{Feature set reduction}, \textit{feature selection}, \textit{prediction performance estimation}, \textit{choice of model complexity} (Supplementary Fig.~S2) and \textit{final model computation} processes were carried out using logistic regression and bootstrap resampling, similarly to the methodology developed in the study of Valli\`eres \textit{et al.} \autocite{vallieres_radiomics_2015} To account for the disproportion of occurrence of events and non-occurrence of events in the training set (15~\% LR, 13~\% DM, 16~\% deaths), an imbalance-adjustment strategy adapted from the study of Schiller \textit{et al.} \autocite{schiller_modeling_2010} was also applied during the training process. Overall for the \textit{PET}, \textit{CT} and \textit{PETCT} feature sets, the number of variables forming the final radiomic models for each outcome were, respectively: I) 8, 3 and 3 radiomic variables for the LR outcome; II) 6, 3 and 3 radiomic variables for the DM outcome; and III) 4, 3 and 6 radiomic variables for the OS outcome. 

The construction of prediction models combining radiomic and clinical variables was then carried out for the nine identified radiomic models (3 feature sets $\times$ 3 outcomes). By estimating prediction performance via stratified random sub-sampling in the training set, the following group of clinical variables were first selected for each outcome: I) \{\textit{Age}, \textit{H}\&\textit{N type}, \textit{T-Stage}, \textit{N-Stage}\} for LR prediction; II) \{\textit{Age}, \textit{H}\&\textit{N type}, \textit{N-Stage}\} for DM prediction; and III) \{\textit{Age}, \textit{H}\&\textit{N type}, \textit{T-Stage}, \textit{N-Stage}\} for OS prediction. Final prediction models were ultimately constructed for each radiomic feature set and H\&N outcome by combining the selected radiomic and clinical variables via random forests and imbalance adjustments.

\subsection*{Performance of prediction models}
The performance of the radiomic prediction models constructed using logistic regression and of the prediction models constructed by combining radiomic and clinical variables via random forests was validated in a testing set consisting of the combination of 106 patients from the H\&N3 and H\&N4 cohorts (Fig.~\ref{fig:MLstrategy}) using receiver operating characteristic (ROC) metrics (binary endpoints). 

\begin{figure}[!htbp] 
	\includegraphics[width=\textwidth]{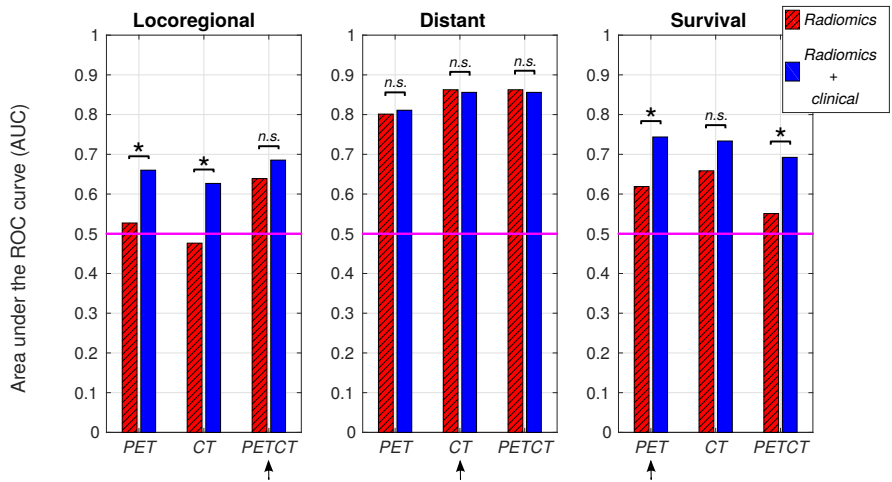}
	\caption{\textbf{Prediction performance of selected models.} \footnotesize{All prediction models were selected and built using the training set (H$\&$N1 and H$\&$N2; $n=194$) for three initial radiomic feature sets: I) PET radiomic features (\textit{PET}); II) CT radiomic features (\textit{CT}); and III) PET and CT radiomic features (\textit{PETCT}). The prediction performance is evaluated here in terms of the area under the receiver operating characteristic curve (AUC) for patients of the testing set (H$\&$N3 and H$\&$N4; $n=106$), for two types of prediction models: I) Radiomic models constructed using logistic regression (\textit{Radiomics}); and  II) Radiomic models combined with clinical variables via random forests (\textit{Radiomics} $+$ \textit{clinical}). Significant increase in AUC from \textit{Radiomics} to \textit{Radiomics} + \textit{clinical} models is identified with an asterisk (*), and non-significant increase is identified by \enquote{\textit{n.s.}}. The radiomic feature sets providing the prediction models with highest performance in this study are identified with an arrow for each outcome.}} \label{fig:PredictPerf}
\end{figure}

Figure~\ref{fig:PredictPerf} presents the performance results (AUC: area under the ROC curve) obtained in the testing set for the \textit{radiomics} and \textit{radiomics} $+$ \textit{clinical} models, where the significance of the increase in AUC when combining clinical to radiomic variables is assessed using the method of DeLong \textit{et al.} \autocite{delong_comparing_1988}. Sensitivity, specificity and accuracy of predictions are also presented in Supplementary Fig.~S3. Overall, it can be observed that there is a general increase in prediction performance for most of the different categories of models that we constructed in this work. For LR prediction, the increase in AUC is significant for prediction models from the \textit{PET} ($p = 0.03$) and the \textit{CT} ($p = 0.01$) radiomic feature sets. For DM prediction, none of the radiomics models show a significant AUC increase when combined with clinical variables. For OS (death) prediction, the increase in AUC is significant for prediction models from the \textit{PET} ($p = 0.01$) and the \textit{PETCT} ($p = 0.006$) radiomic feature sets. Furthermore, we verified that the increase in performance  is not explained by the use of a more complex and potentially more predictive learning algorithm: random forests classifiers constructed with radiomic variables alone preserved the predictive properties obtained by logistic regression models constructed with the same variables, but without improving them (Supplementary Table~S3). These results point to the potential of random forests in successfully combining the complementary value of different categories of prognostic factors such as radiomic and clinical variables.

In Fig.~\ref{fig:PredictPerf}, the highest performance for LR prediction was obtained using the model combining the \textit{PETCT} radiomic and clinical variables, with an AUC of 0.69. For DM prediction, the highest performance was obtained using the \textit{CT} radiomic model, with an AUC of 0.86. These results demonstrate that different radiomic-based models could successfully be used to predict specific outcomes such as locoregional recurrences and distant metastases in H\&N cancer. Finally, the highest performance for OS (death) prediction was obtained using the model combining the \textit{PET} radiomic and clinical variables, with an AUC of 0.74. For subsequent analysis in the next section, only the prediction models (\textit{radiomics} and \textit{radiomics} $+$ \textit{clinical}) constructed from these radiomic features sets (\textit{PETCT} for LR, \textit{CT} for DM, \textit{PET} for OS) are used. The complete description of these identified radiomic  models (specific features, texture extraction parameters, logistic regression coefficients) is given in Supplementary Results
section 1.4.2.

\subsection*{Comparison with other prognostic factors}
The performance of the best radiomic prediction models and the best prediction models combining radiomic and clinical variables identified in this study (shown with arrows in Fig.~\ref{fig:PredictPerf}) were further compared against other prognostic factors: I) \textit{Volume}; II) \enquote{clinical-only} models; III) combination of \textit{Volume} and clinical variables; and IV) a validated radiomic signature developed for the prognosis assessment of overall survival \autocite{aerts_decoding_2014,leijenaar_external_2015}. In addition to the prediction performance evaluated using ROC metrics, the prognostic performance of the models was also assessed using: I) the concordance index (CI) \autocite{harrell_multivariable_1996} between the output probability of occurrence of an event (LR, DM, death) of prediction models and the time elapsed before an event occurred (\enquote{time-to-event}); and II) the \textit{p}-value obtained from Kaplan-Meier analysis using the log-rank test between two risk groups. The models consisting of only radiomic or the \textit{Volume} variables were optimized using logistic or cox regression, and all models involving clinical variables were optimized using random forest classifiers, still using the defined training set of this work (H\&N1 and H\&N2 cohorts; $n = 194$). Fully independent results are then presented in Table~\ref{tab:comparison} for models evaluated in the testing set (H\&N3 and H\&N4 cohorts; $n = 106$).

For locoregional recurrences, we found that the model combining the \textit{PETCT} radiomic and clinical variables provided the best performance in terms of predictive/prognostic power and balance of classification of occurrence of events and non-occurrence of events, notably with an AUC of 0.69, a sensitivity of 0.63, a specificity of 0.68, an accuracy of 0.67, a CI of 0.67 and a Kaplan-Meier \textit{p}-value of 0.03. Using random permutation tests, each variable was calculated to be approximately of equal importance in the random forest model (Supplementary Table~S4). Similarly to univariate analysis, \textit{Volume} was not found to be a significant prognostic factor for LR. On the other hand, clinical variables alone had high performance with an AUC of 0.72 and a CI of 0.69, but this type of modeling did not provide sufficient balance between the prediction of occurrence and non-occurrence of events (sensitivity of 0.50, specificity of 0.76).

For distant metastases, we found that the model combining the \textit{CT} radiomic and clinical variables provided the best overall performance, notably with an AUC of 0.86, a sensitivity of 0.86, a specificity of 0.76, an accuracy of 0.77, a CI of 0.88 and a Kaplan-Meier \textit{p}-value of $3 \times 10^{-6}$. However, radiomic variables were found to be of much higher importance than the clinical variables in the random forest model (Supplementary Table~S4). In fact, the model composed of clinical variables alone did not perform well. \textit{Volume} was again found to be a significant prognostic factor for DM, but radiomic variables outperformed it.

For overall survival, we found that the model composed of clinical variables alone provided the best overall performance, notably with an AUC of 0.78, a sensitivity of 0.92, a specificity of 0.57, an accuracy of 0.65, a CI of 0.76 and a Kaplan-Meier \textit{p}-value of $3 \times 10^{-5}$. Furthermore, the \textit{H}\&\textit{N type} variable had the highest and \textit{N-Stage} the lowest importance in the model (Supplementary Table~S4). Another important finding was that \textit{Volume} alone provided similar or better prognosis assessment of OS than any of the following radiomic-based models: I) the best radiomic model for OS constructed in this work; II) the original radiomic signature using the cox regression coefficients employed in the work of Aerts \& Velazquez \textit{et al.} \autocite{aerts_decoding_2014}; and  III) a revised version of the radiomic signature computation (Supplementary Section 2.6.2) using new sets of regression coefficients trained with the current training set of this work.

\subsection*{Risk assessment of tumour outcomes}
\begin{figure}[!htbp] 
	\includegraphics[width=\textwidth]{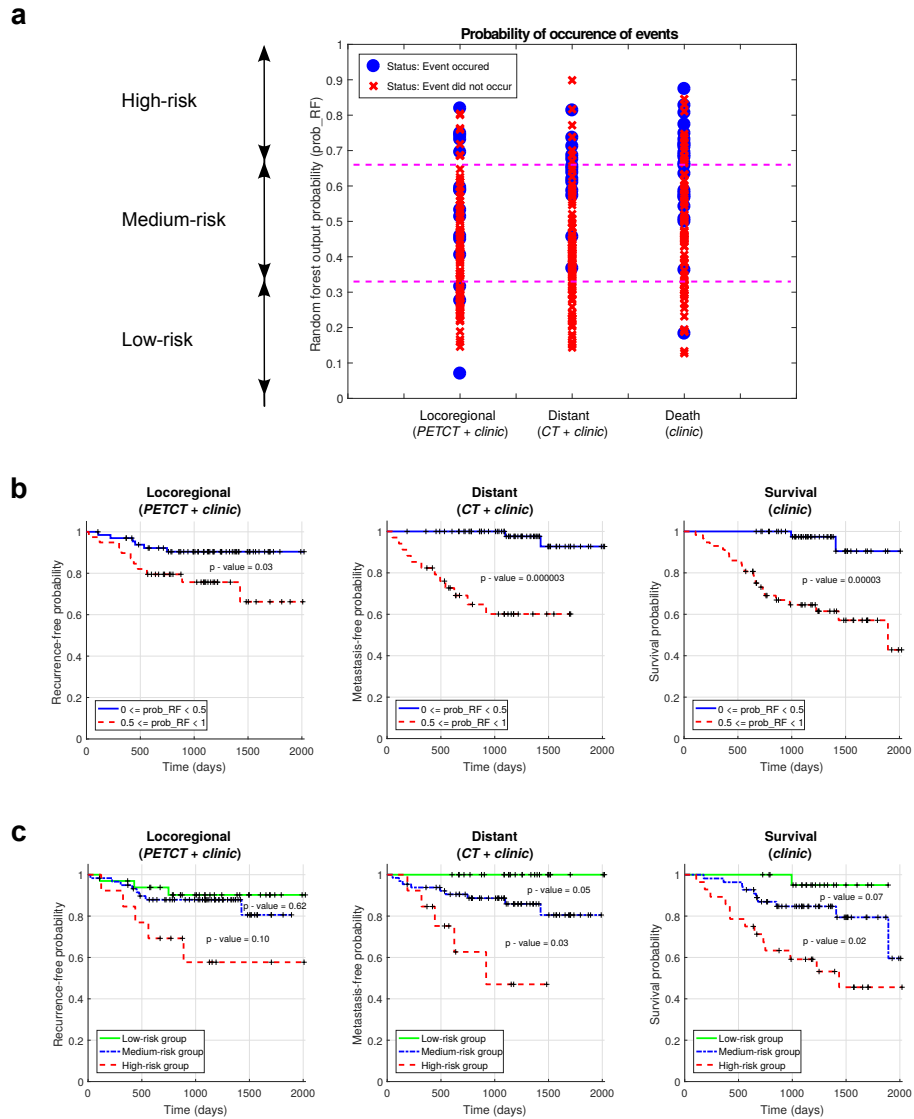}
	\caption{\textbf{Risk assessment of tumour outcomes.} \footnotesize{\textbf{(a)} Probability of occurrence of events (locoregional recurrence, distant metastases, death) for each patient of the testing set (H$\&$N3 and H$\&$N4; $n=106$) as determined by the random forest classifiers built using the training set (H$\&$N1 and H$\&$N2; $n=194$). The output probability of occurrence of events of random forests allows for risk stratification; for example, three risk groups can be defined (low, medium, high) using probability thresholds of $\frac{1}{3}$ and $\frac{2}{3}$. \textbf{(b)} Kaplan-Meier curves of the testing set using a risk stratification into two groups as defined by a random forest output probability threshold of 0.5. All curves have significant prognostic performance, thus demonstrating the possibility of outcome-specific risk assessment in head-and-neck cancer. \textbf{(c)} Kaplan-Meier curves of the testing set using a risk stratification in three groups as defined by random forest output probability thresholds of $\frac{1}{3}$ and $\frac{2}{3}$. Some pair of curves have significant prognostic performance, thus demonstrating the possibility of risk stratification into multiple groups for treatment escalation/personalization in head- and-neck cancer.}} \label{fig:RiskAssessment}
\end{figure}

The work performed in this study leads to the identification of three prediction models based on three final random forest classifiers, one for each of the outcome studied here (identified with italic fonts in Table~\ref{tab:comparison}): I) \{\textit{PET-GLN}$_{\mathrm{GLSZM}}$, \textit{CT-Correlation}$_{\mathrm{GLCM}}$, \textit{CT-LGZE}$_{\mathrm{GLSZM}}$, \textit{age}, \textit{H}\&\textit{N type}, \textit{T-Stage}, \textit{N-Stage}\} for LR; II) \{\textit{CT-LRHGE}$_{\mathrm{GLRLM}}$, \textit{CT-ZSV}$_{\mathrm{GLSZM}}$, \textit{CT-ZSN}$_{\mathrm{GLSZM}}$, \textit{age}, \textit{H}\&\textit{N type}, \textit{N-Stage}\} for DM; and III) \{\textit{age}, \textit{H}\&\textit{N type}, \textit{T-Stage}, \textit{N-Stage}\} for OS. A property of a random forest is that the binary prediction of each of its decision tree can be averaged to serve as an output probability of occurrence of a given event (\textit{prob}$_{\mathrm{RF}}$). This output probability, similarly to other machine learning algorithms, can constitute one of the tools to be used for the risk assessment of specific tumour outcomes. For example, the final random forest classifiers constructed in the training set (H\&N1 and H\&N2 cohorts; $n = 194$) can be used to stratify the risk of occurrence of the outcome events for each patient of the testing set (H\&N3 and H\&N4 cohorts; $n = 106$) into three groups (Fig.~\ref{fig:RiskAssessment}a): I) low-risk group $\rightarrow 0 \leq \mathit{prob}_{\mathrm{RF}} < \frac{1}{3}$; II) medium-risk group $\rightarrow \frac{1}{3} \leq \mathit{prob}_{\mathrm{RF}} < \frac{2}{3}$; and III) high-risk group $\rightarrow \frac{2}{3} \leq \mathit{prob}_{\mathrm{RF}} < 1$. Thereafter, this stratification scheme can be used to evaluate the probability of non-occurrence of the events after a given time for the different risk groups via Kaplan-Meier analysis. Standard Kaplan-Meier analysis using two risk groups ($\mathit{prob}_{\mathrm{RF}} \leq 0.5$, $\mathit{prob}_{\mathrm{RF}} > 0.5$) is first shown in Fig.~\ref{fig:RiskAssessment}b for all patients of the testing set. These curves demonstrate the possibility of prognostic risk assessment of specific outcomes in H\&N cancer such as locoregional recurrences ($p = 0.03$) and distant metastases ($p = 3 \times 10^{-6}$) using specific prediction models combining different radiomic and clinical variables, but also of the general outcome of overall survival ($p = 3 \times 10^{-5}$) using a prediction model composed of clinical variables only. More accurate prognostic risk assessment can then be further performed using Kaplan-Meier analysis with three risk groups (as defined above: low-risk, medium-risk, high-risk) as shown in Fig.~\ref{fig:RiskAssessment}c for all patients of the testing set. For the risk assessment of LR, the developed prediction model is, however, not powerful enough to significantly separate the patients between the high/medium ($p = 0.62$) and medium/low ($p = 0.10$) risk groups. In the case of DM, the developed prediction model allows to significantly separate the patients between the high/medium ($p = 0.05$) and medium/low ($p = 0.03$) risk groups. For OS, the developed prediction model does not significantly separate the patients between the high/medium risk groups ($p = 0.07$), but it does significantly separate the patients between the medium/low risk groups ($p = 0.02$).

\singlespacing
\begin{table}[!htbp]
\caption{\textbf{Comparison of prediction/prognostic performance of models constructed in this work with other variable combinations.} Performance is shown for models constructed in the training set (H$\&$N1 and H$\&$N2; $n=194$) and independently evaluated in the testing set (H$\&$N3 and H$\&$N4; $n=106$).} \label{tab:comparison}
\noindent\makebox[\textwidth]{ 
\small
\begin{tabular}{c | c | *{4}{c} | *{2}{c}}

\hline
\multirow{2}{*}{Outcome} & \multirow{2}{*}{Variables} & \multicolumn{4}{|c}{Prediction} & \multicolumn{2}{|c}{Prognosis} \\
\cline{3-8}
& & AUC$^\mathrm{a}$ & Sensitivity$^\mathrm{a}$ & Specificity$^\mathrm{a}$ & Accuracy$^\mathrm{a}$ & CI$^\mathrm{b}$ & \textit{p}-value$^\mathrm{c}$ \\
\hline
\hline

\multirow{5}{*}{Locoregional}  & $\mathrm{Radiomics_{PETCT}}$                       & 0.64 & 0.56 & 0.67 & 0.65 & 0.63 & 0.28 \\
			                  & $\mathrm{Volume}$                                & 0.43 & 0.31 & 0.58 & 0.54 & 0.40 & 0.80 \\
			                  & $\mathrm{Clinical}$                              & 0.72 & 0.50 & 0.76 & 0.72 & 0.69 & 0.05 \\
			                  & $Radiomics_{PETCT} + Clinical$   & \emph{0.69} & \emph{0.63} & \emph{0.68} & \emph{0.67} & \emph{0.67} & \emph{0.03} \\
			                  & $\mathrm{Volume} + \mathrm{Clinical}$            & 0.71 & 0.50 & 0.76 & 0.72 & 0.68 & 0.06 \\
\hline

\multirow{5}{*}{Distant}        & $\mathrm{Radiomics_{CT}}$                        & 0.86 & 0.79 & 0.77 & 0.77 & 0.88 & 0.0001 \\
			                  & $\mathrm{Volume}$                                & 0.80 & 0.86 & 0.65 & 0.68 & 0.83 & 0.10 \\
			                  & $\mathrm{Clinical}$                              & 0.55 & 0.64 & 0.46 & 0.48 & 0.60 & 0.61 \\
			                  & $Radiomics_{CT} + Clinical$    & \emph{0.86} & \emph{0.86} & \emph{0.76} & \emph{0.77} & \emph{0.88} & \emph{0.000003} \\
			                  & $\mathrm{Volume} + \mathrm{Clinical}$            & 0.78 & 1 & 0.50 & 0.57 & 0.80 & 0.0004 \\
\hline

\multirow{5}{*}{Survival}       & $\mathrm{Radiomics_{PET}}$                        & 0.62 & 0.58 & 0.66 & 0.64 & 0.60 & 0.03 \\
			                  & $\mathrm{Volume}$                                & 0.68 & 0.67 & 0.57 & 0.59 & 0.67 & 0.29 \\
			                  & $Clinical$                              & \emph{0.78} & \emph{0.92} & \emph{0.57} & \emph{0.65} & \emph{0.76} & \emph{0.00003} \\
			                  & $\mathrm{Radiomics_{PET}} + \mathrm{Clinical}$    & 0.74 & 0.79 & 0.57 & 0.62 & 0.71 & 0.002 \\
			                  & $\mathrm{Volume} + \mathrm{Clinical}$            & 0.79 & 0.88 & 0.52 & 0.60 & 0.76 & 0.0006 \\
\hline

\multirow{3}{*}{Survival$^{\mathrm{d}}$}     & $\mathrm{{Radiomics_{CTcompleteSign}}^e}$              & -- & -- & -- & -- & 0.66 & 0.70 \\
											& $\mathrm{{Radiomics_{CTsign}}^f}$                      & 0.68 & 0.71 & 0.50 & 0.55 & 0.66 & 0.05 \\
                                             & $\mathrm{{Radiomics_{CTsign}}^g} + \mathrm{Clinical}$  & 0.80 & 0.96 & 0.38 & 0.51 & 0.75 & 0.001 \\
\hline

\end{tabular}}
\begin{flushleft}
\scriptsize
\item[] $\rightarrow$ Models involving \textit{Radiomic} variables only or the \textit{Volume} variable only were optimized using logistic/cox regression. All models involving \textit{Clinical} variables were optimized using random forests. 
\item[] $\rightarrow$ The best predictive/prognostic and balanced models for each outcome (final models) are identified in italic and are fully described in Supplementary Table~S4.
\item[] $^\mathrm{a}$ Binary prediction of outcome using logistic regression/random forest output responses.
\item[] $^\mathrm{b}$ Concordance-index between cox regression/random forest output responses and time to events.
\item[] $^\mathrm{c}$ Log-rank test from Kaplan-Meier curves with a risk stratification into two groups (thresholds: median hazard ratio for cox regression, output probability of 0.5 for random forests).
\item[] $^\mathrm{d}$  Radiomic signature variables as defined in Aerts \& Velazquez \textit{et al.} \autocite{aerts_decoding_2014}
\item[] $^\mathrm{e}$ Using the original definition of the radiomic signature variables, and the original cox regression coefficients and median hazard ratio trained from the Lung1 cohort in the study of Aerts \& Velazquez \textit{et al.} \autocite{aerts_decoding_2014}
\item[] $^\mathrm{f}$ Using a revised version of the radiomic signature variables (Supplementary Methods section 2.6.2) and new cox/logistic regression coefficients trained using the current training set of this work.
\item[] $^\mathrm{g}$ Using a revised version of the radiomic signature variables (Supplementary Methods section 2.6.2) and a random forest classifer trained using the current training set of this work.
\end{flushleft}
\end{table}

\onehalfspacing
\newpage

\section*{Discussion}
Increasing evidence suggests that the genomic heterogeneity of aggressive tumours could translate into intratumoural spatial heterogeneity exhibited at the anatomical and functional scales \autocite{segal_decoding_2007,diehn_identification_2008,aerts_decoding_2014}. This constitutes the central idea of the emerging field of \enquote{radiomics}, in which large amounts of information via advanced quantitative analysis of medical images are used as non-invasive means to characterize intratumoural heterogeneity and to reveal important prognostic information about the cancer \autocite{el_naqa_exploring_2009,gillies_biology_2010,lambin_radiomics:_2012,kumar_radiomics:_2012}. Ultimately, the objective is to narrow down this extensive quantity of information into simple prediction models that can aid in the identification of specific tumour phenotypes for improved treatment management. In this study, we were able via advanced machine learning to develop two prediction models combining PET/CT radiomics and clinical information for the early assessment of the risk of locoregional recurrences and distant metastases in head-and-neck cancers.

First, we extracted a total of 1615 radiomic features from PET and CT pre-treatment images of 300 patients with head-and-neck cancer from four different cohorts. These features are composed of 10 intensity features, 5 shape features and 40 textures computed using 40 different combinations of extraction parameters (five isotropic voxel sizes, two quantization algorithms and four numbers of gray levels). In general, different texture features will better represent the underlying tumour biology using different extraction parameters, and the optimal set of parameters to use is application-specific and depends on many factors such as the clinical endpoint studied and the imaging modalities employed. Texture optimization has the potential to enhance the predictive value of the extracted features as Valli\`eres \textit{et al.} \autocite{vallieres_radiomics_2015} have previously shown, and we suggest to incorporate this step in the texture extraction workflow of future similar studies.

Univariate analysis showed that the majority of the features extracted from both PET and CT images are significantly associated with the development of distant metastases, suggesting that the metastatic phenotype of tumours can be captured via quantitative image analysis. On the other hand, none of the radiomic features were significantly associated with locoregional recurrences after multiple testing corrections with a FDR of 10~\%. Although combinations of these metrics still proved useful for prognostic risk assessment, it does reveal the need of using other types of metrics such as radiation dose characteristics to enhance the predictive properties of the models constructed for locoregional recurrences. In addition to radiomic features, we also investigated the association of clinical variables with the different head-and-neck cancer outcomes studied in this work. The most significant association was found between \textit{HPV status} and locoregional recurrences, a currently known result that agrees with other studies \autocite{fakhry_improved_2008,ang_human_2010}. However, this result was obtained with only 120 of the 300 patients with available \textit{HPV status}, and this variable could not be used in the subsequent multivariable analysis.

Next, we constructed multivariable prediction models from radiomic information alone by using the methodology developed by Valli\`eres \textit{et al.} \autocite{vallieres_radiomics_2015} All models were entirely produced from the defined training set of this work combining two head-and-neck cancer cohorts (H\&N1 and H\&N2; $n = 194$). The best radiomic model for locoregional recurrences (Table~\ref{tab:comparison}) was found to possess good predictive properties in the defined testing set of this work combining two head-and-neck cancer cohorts (H\&N3 and H\&N4; $n = 106$). This model is composed of one metric extracted from PET images (\textit{GLN}$_{\mathrm{GLSZM}}$: \textit{gray-level nonuniformity}$_{\mathrm{GLSZM}}$) and two metrics extracted from CT images (\textit{correlation}$_{\mathrm{GLCM}}$ and \textit{LGZE}$_{\mathrm{GLSZM}}$: \textit{low gray-level zone emphasis}$_{\mathrm{GLSZM}}$). The best radiomic model for distant metastases (Table~\ref{tab:comparison}) was found to possess high predictive properties in the testing set, and is composed of three metrics extracted from CT images (\textit{LRHGE}$_{\mathrm{GLRLM}}$: \textit{long run high gray level emphasis}$_{\mathrm{GLRLM}}$, \textit{ZSV}$_{\mathrm{GLSZM}}$: \textit{zone size variance}$_{\mathrm{GLSZM}}$ and \textit{ZSN}$_{\mathrm{GLSZM}}$: \textit{zone size nonuifomity}$_{\mathrm{GLSZM}}$). These results suggest that radiomic models can be specific enough to assess the risk of different outcomes in head-and-neck cancer. The models we developed for locoregional recurrences and distant metastases are in fact overall different and they capture specific tumour phenotypes. It is also noteworthy that all the selected radiomic features of these two models are textural, attesting to the high potential of textures to characterize the complexity of spatial patterns within tumours. As mentioned earlier, aggressive tumours tend to show increased intratumoural heterogeneity \autocite{fidler_critical_1990,yokota_tumor_2000,campbell_patterns_2010}, notably in terms of the heterogeneity in size and intensity characteristics of the different tumour sub-regions in PET and CT images. This effect may be captured by the \textit{PET-GLN}$_{\mathrm{GLSZM}}$, \textit{CT-LRHGE}$_{\mathrm{GLRLM}}$, \textit{CT-ZSV}$_{\mathrm{GLSZM}}$ and \textit{CT-ZSN}$_{\mathrm{GLSZM}}$ texture features in our radiomic models, a result in agreement with a previous study describing the importance of zone-size nonuniformities for the prognostic assessment of head-and-neck tumours \autocite{cheng_zone-size_2014}. From our experience, we have observed that aggressive tumours also frequently contain large inactive or necrotic regions of uniform intensities, suggesting that these tumours could be rapidly increasing in size and that they could be more at risk to metastasize, for example \autocite{vakkila_inflammation_2004,proskuryakov_mechanisms_2010,ahn_necrotic_2016}. Here, this effect may be captured by the \textit{CT-Correlation}$_{\mathrm{GLCM}}$ and \textit{CT-LGZE}$_{\mathrm{GLSZM}}$ texture features. Overall, these results suggest that radiomic features could be useful to improve our understanding of the underlying biology of tumours.

We also attempted in this study to improve the predictive power of our prediction models by combining radiomic variables with clinical data. The first step of our method is based on a fast mining of radiomic variables using logistic regression. Then, random forests \autocite{breiman_random_2001} are used as a means to combine radiomic and clinical information into a single classifier. It would also be feasible to only use random forests to mine the radiomic variables, but our method is advantageous in terms of computation speed. Our results showed that the combination of clinical variables with the optimal radiomic variables via random forests had a positive impact on the prediction and prognostic assessment of locoregional recurrences and distant metastases, although with minimal impact in the latter case (Fig.~\ref{fig:PredictPerf}, Table~\ref{tab:comparison}). As seen in Supplementary Table~S4, this can be explained from the fact that the identified radiomic features are the strong and dominant variables in the model for distant metastases predictions. Nonetheless, we believe that random forests is one effective algorithm well-suited to combine variables of different types (categorical and continuous inputs) such as clinical and panomic tumour profile information. In general, the ongoing optimization of machine learning techniques in radiomic applications \autocite{parmar_radiomic_2015-1,parmar_machine_2015,parmar_radiomic_2015} is a step forward to improve clinical predictions.

In this work, we also performed a comprehensive comparison of the prediction/prognostic performance of radiomics versus clinical models and volumetric variables (Table~\ref{tab:comparison}). Metabolic tumour volume has already been shown to be an independent predictor of outcomes in head-and-neck cancer \autocite{tang_validation_2012}, but it was also suggested by Hatt et al. \autocite{hatt_18f-fdg_2015} that heterogeneity quantification via texture analysis may provide valuable complementary information to the tumour volume variable for volumes above 10~cm$^3$. In this study, 85~\% of the patients had a gross tumour volume greater than 10~cm$^3$ and we consequently found that radiomic models performed considerably better than tumour volume alone for the prediction of locoregional recurrences and distant metastases. On the other hand, clinical variables alone did not perform well on their own for distant metastases, but they had good performance for locoregional recurrences by outperforming radiomic models, thus suggesting that our radiomic models need to be improved to better model locoregional recurrences.

In terms of overall survival assessment, our results indicate that the tumour volume variable matched or outperformed all radiomic models thus far we developed or tested in this work, including a previously validated radiomic signature \autocite{aerts_decoding_2014,leijenaar_external_2015}. For one, it is unsurprising that the original radiomic signature \autocite{aerts_decoding_2014} did not perform better than tumour volume, as it can be verified that all its feature components are very strongly correlated with tumour volume: the Pearson linear coefficients between tumour volume and the four features of the signature \autocite{aerts_decoding_2014} were calculated to be 0.62 (\textit{Energy}), 0.80 (\textit{Compactness}), 0.99 (\textit{GLN}$_{\mathrm{GLRLM}}$) and 0.94 (\textit{GLN}$_{\mathrm{GLRLM}}$\textit{\_HLH}) using the whole set of 300 patients of this study, all with $p \ll 0.001$. On the other hand, all the features forming the other radiomic models developed in this work showed potential complementarity value to tumour volume (but the models still did not perform better than tumour volume alone for overall survival assessment): all the features of the revised version of the radiomic signature (Supplementary Section 2.6.2) had a Pearson linear coefficient lower than 0.5 except one (\textit{Energy}), and all the variables forming the final radiomic models constructed in this work (italic fonts in Table~\ref{tab:comparison}, including those for locoregional recurrences and distant metastases) had linear coefficients lower than 0.40. This suggests that overall survival may be harder to model than specific tumour outcomes due to a larger number of confounding factors being involved, and it may thus be more prone to overfitting during training. As a consequence, tumour volume may currently be a more robust and reproducible metric than imaging features for modeling this outcome.  In the end, the best global performance for overall survival was however obtained with clinical variables alone. This would emphasize that clinical data remains the important source of information to consider for the evaluation of the likelihood of occurrence of a general outcome with many confounding factors such as overall survival, and that more work is required to understand how to adequately model overall survival using radiomic features.

The optimal results in terms of predictive/prognostic performance and balance of prediction between the occurrence and non-occurrence of locoregional recurrences and distant metastases were found in this work by constructing models combining radiomic and clinical variables via random forests (full description of the models in Supplementary Results). Compared to the general assessment of overall survival as in previous studies \autocite{aerts_decoding_2014,leijenaar_external_2015}, our results demonstrate the possibility of decoding specific tumour phenotypes for the risk assessment of specific outcomes in head-and-neck cancer. The final results obtained for distant metastases were considerably higher than those obtained for overall survival, but those obtained for locoregional recurrences were lower albeit clinically significant (Table~\ref{tab:comparison}). Also, as seen in Fig.~\ref{fig:RiskAssessment} with patients of the testing set, the output probability of occurrence of events of our prediction models allow to significantly separate patients into two locoregional recurrence risk groups and into three distant metastases risk groups. The clinical impact of our results and of the risk assessment of specific outcomes in head-and-neck cancer could be substantial, as it could allow for a better personalization of treatments. For example, higher radiation doses could be considered for patients at higher risks of locoregional recurrences. For distant metastases, the chemotherapy regimens could be strengthened for patients in the high risk group to reduce potential metastatic invasion, and lessened for patients in the low risk group to improve quality of life. These are hypothetical scenarios that, at the moment, are not ready to be implemented in the clinical environment, as our models first need to be constructed and validated on larger patient cohorts, and robust clinical trials are required to validate their benefits on patient survival. Furthermore, the heterogeneity of the patient cohorts used in this work including varying image acquisition parameters may undermine the power of the developed models. However, it may also improve their generalizability, and the results presented in this study could now be useful for the generation of new hypotheses driving future prospective studies.

Overall, we showed in this study that radiomics provide important prognostic information for the risk assessment of locoregional recurrences and distant metastases in head-and-neck cancer. In general, the combination of panomics data into clinically-integrated prediction models should allow to more comprehensively assess cancer risks and could improve how we adapt treatments for each patient. As the standardization efforts of radiomics analysis continue to rapidly progress \autocite{nyflot_quantitative_2015,zhao_reproducibility_2016}, we can envision the clinical implementation of radiomics-based decision-support systems in the future. Full transparency on data and methods is the key for the progression of the field, and our research efforts needs to include large-scale collaborations and reproducibility practices to increase the possibility of translation of radiomics into the clinical environment \autocite{ioannidis_how_2014}.
\newpage

\section*{Methods}

\subsection*{Data sets availability}
Our analysis was conducted on imaging and clinical data of a total of 300 H\&N cancer patients from four different institutions who received radiation alone ($n = 48$, 16~\%) or chemo-radiation ($n = 252$, 84~\%) with curative intent as part of treatment management. The median follow-up period of all patients was 43 months (range: 6-112). The Institutional Review Boards of all participating institutions approved the study. Retrospective analyses were performed in accordance with the relevant guidelines and regulations as approved by the Research Ethics Committee of McGill University Health Center (Protocol Number: MM-JGH-CR15-50).

\begin{itemize}
\item The H\&N1 data set consists of 92 head-and-neck squamous cell carcinoma (HNSCC) patients treated at H\^opital g\'en\'eral juif (HGJ) de Montr\'eal, QC, Canada. During the follow-up period, 12 patients developed a locoregional recurrence (13~\%), 16 patients developed distant metastases (17~\%) and 14 patients died (15~\%). This data set was used as part of the \emph{training set} of this work.
\item The H\&N2 data set consists of 102 head-and-neck squamous cell carcinoma (HNSCC) patients treated at Centre hospitalier universitaire de Sherbrooke (CHUS), QC, Canada. During the follow-up period, 17 patients developed a locoregional recurrence (17~\%), 10 patients developed distant metastases (10~\%) and 18 patients died (18~\%). This data set was used as part of the \emph{training set} of this work.
\item The H\&N3 data set consists of 41 head-and-neck squamous cell carcinoma (HNSCC) patients treated at H\^opital Maisonneuve-Rosemont (HMR) de Montr\'eal, QC, Canada. During the follow-up period, 9 patients developed a locoregional recurrence (22~\%), 11 patients developed distant metastases (27~\%) and 19 patients died (46~\%). This data set was used as part of the \emph{testing set} of this work.
\item The H\&N4 data set consists of 65 head-and-neck squamous cell carcinoma (HNSCC) patients treated at Centre hospitalier de l'Universit\'e de Montr\'eal (CHUM), QC, Canada. During the follow-up period, 7 patients developed a locoregional recurrence (11~\%), 3 patients developed distant metastases (5~\%) and 5 patients died (8~\%). This data set was used as part of the \emph{testing set} of this work.
\end{itemize}

All patients underwent FDG-PET/CT imaging scans within a median of 18 days (range: 6-66) before treatment. For 93 of the 300 patients (31~\%), the radiotherapy contours were directly drawn on the CT of the PET/CT scan by expert radiation oncologists and thereafter used for treatment planning. For 207 of the 300 patients (69~\%), the radiotherapy contours were drawn on a different CT scan dedicated to treatment planning and were propagated/resampled to the FDG-PET/CT scan reference frame using intensity-based free-form deformable registration with the software MIM$^{\mathrm{\textregistered}}$ (MIM software Inc., Cleveland, OH).

Further information specific to each patient cohort (e.g. treatment details) is presented in Supplementary Methods section 2.4 and Supplementary Tables~S5-S8. Pre-treatment FDG-PET/CT imaging data, clinical data, radiotherapy contours (\textit{RTstruct}) and MATLAB$^{\mathrm{\textregistered}}$ routines allowing to read imaging data and their associated region-of-interest (ROI) are made available for all patients on The Cancer Imaging Archive (TCIA) \autocite{clark_cancer_2013} at: \url{http://www.cancerimagingarchive.net}. The Research Ethics Committee of McGill University Health Center approved online publishing of clinical and imaging data following patient anonymisation.

\subsection*{Sample size and division of cohorts}
Patients with recurrent H\&N cancer or with metastases at presentation, and patients receiving palliative treatment were excluded from the study. Patients that did not develop a locoregional recurrence or distant metastases during the follow-up period and that had a follow-up time smaller than 24 months were also excluded from the study. The four patient cohorts were then divided into two groups to create one combined training set (H\&N1 and H\&N2; $n = 194$) and one combined testing set (H\&N3 and H\&N4; $n = 106$). Bootstrap resampling and stratified random sub-sampling were always performed with patients from the training set to estimate the relevant performance metrics of interest and to construct the final prediction models, and fully independent validation results were computed with patients from the testing set. This precise type of division of patient cohorts allowed to: I) Train on a combined set of different cohorts to allow the models to take into account some institutional variability; II) Reduce the number of testing results reported; III) Create a training set size to testing set size ratio of approximately 2:1; and IV) Conduct partition sampling such that the proportion of occurrence of events (locoregional recurrences, distant metastases) are approximately the same in the training and testing sets.

\subsection*{Extraction of radiomic features}
Starting from the original FDG-PET/CT imaging data and associated radiotherapy contours in DICOM format, the complete set of data was read and transferred into MATLAB$^{\mathrm{\textregistered}}$ (MathWorks, Natick, MA) format using in-house routines. PET images were converted to standard uptake value (SUV) maps and CT images were kept in raw Hounsfield Unit (HU) format. In this work, we then extracted a total of 1615 radiomic features for both the PET and CT images from the tumour region defined by the \enquote{GTV$_{\mathrm{primary}}$ + GTV$_{\mathrm{lymph\ nodes}}$} contours as delineated by the radiation oncologists of each institution. These features can be divided into three different groups: I) 10 first-order statistics features (intensity); II) 5 morphological features (shape); and III) 40 texture features each computed using 40 different combinations of extraction parameters.

Intensity features are computed from histograms ($n_{bins} = 100$) of the intensity distribution of the ROI. The features extracted in this work were the \textit{variance}, the \textit{skewness}, the \textit{kurtosis}, \textit{SUVmax}, \textit{SUVpeak}, \textit{SUVmean}, the \textit{area under the curve of the cumulative SUV-volume histogram} \autocite{van_velden_evaluation_2011}, the \textit{total lesion glycolysis}, the \textit{percentage of inactive volume} and the \textit{generalized effective total uptake} \autocite{rahmim_novel_2016}. Shape feature describe geometrical aspects of the ROI. The features extracted in this work were the \textit{volume}, the \textit{size} (maximum tumour diameter), the \textit{solidity}, the \textit{eccentricity} and the \textit{compactness}.

Texture features measure intratumoural heterogeneity by quantitatively describing the spatial distributions of the different intensities within the ROI. In this work, 9 features from the Gray-Level Co-occurrence Matrix (GLCM) \autocite{haralick_textural_1973}, 13 features from the Gray-Level Run-Length Matrix (GLRLM) \autocite{galloway_texture_1975,chu_use_1990,dasarathy_image_1991}, 13 features from the Gray-Level Size Zone Matrix (GLSZM) \autocite{galloway_texture_1975,chu_use_1990,dasarathy_image_1991,thibault_texture_2009} and 5 features from the Neighbourhood Gray-Tone Difference Matrix (NGTDM) \autocite{amadasun_textural_1989} were computed. All texture matrices were constructed using 3D analysis/26-voxel connectivity of the tumour region resampled to a defined isotropic voxel size. For each of the four texture types, only one matrix was computed per scan by simultaneously taking into account the neighbouring properties of voxels in the 13 directions of 3D space. However, the 6 voxels at a distance of 1 voxel, the 12 voxels at a distance of $\sqrt{2}$ voxels, and the 8 voxels at a distance of $\sqrt{3}$ voxels around center voxels were treated differently in the calculation of the matrices to take into account discretization length differences.

All 40 texture features from the ROI of both PET and CT volumes were extracted using all possible combinations (40) of the following parameters:

\begin{itemize}
\item Isotropic voxel size (5): Voxel sizes of 1~mm, 2~mm, 3~mm, 4~mm and 5~mm.
\item Quantization algorithm (2): \textit{Equal-probability} (equalization of intensity histogram) and \textit{Uniform} (uniform division of intensity range) quantization algorithms with fixed number of gray levels.
\item Number of gray levels (4): Fixed number of gray levels of 8, 16, 32 and 64 in the quantized ROI.
\end{itemize}

Detailed description with supplementary references and methodology used to extract all radiomic features is further provided in Supplementary Methods section 2.5.

\subsection*{Construction of radiomic models}
The construction of prediction models from the total set of radiomic features for each of the three initial feature sets (I: PET features; II: CT features; and III: PET and CT features) and three H\&N cancer outcomes was performed from the defined training set of this work (H\&N1 and H\&N2 cohorts; $n = 194$) using the methodology developed in the work of Valli\`eres \textit{et al.} \autocite{vallieres_radiomics_2015} The process of combining radiomic features into a multivariable model was achieved using the logistic regression utilities of the software DREES \autocite{el_naqa_dose_2006}. The general workflow is presented in Supplementary Fig.~S5.

First, feature set reduction was performed for each of the initial feature sets via a stepwise forward feature selection scheme in order to create reduced feature sets containing 25 different features balanced between predictive power (Spearman's rank correlation) and non-redundancy (maximal information coefficient\autocite{reshef_detecting_2011}). This procedure was carried out using the \textit{Gain} equation\autocite{vallieres_radiomics_2015}, which is fully detailed in Supplementary Methods section 2.2.2.

From the reduced feature sets, stepwise forward feature selection was then carried out by maximizing the 0.632+ bootstrap AUC \autocite{efron_improvements_1997,sahiner_classifier_2008}. For a given model order (number of combined variables) and a given reduced feature set, the feature selection step was divided into 25 experiments. In each of these experiments, all the different features from the reduced set were used as different \enquote{starters}. For a given starting feature, 100 logistic regression models or order 2 were first created using bootstrap resampling (100 samples) for each of the remaining features in the reduced feature set. Then, the single remaining feature that maximized the 0.632+ bootstrap AUC of the 100 models was chosen, and the process was repeated up to model order 10. Finally, for each model order of each feature set, the experiment that yielded the highest 0.632+ bootstrap AUC was identified, and combinations of features were chosen for model orders of 1 to 10. The whole feature selection process is pictured in Supplementary Fig.~S6.

Once optimal combinations of features were identified for model orders of 1 to 10 for all feature sets, prediction performance was estimated using the 0.632+ bootstrap AUC (100 samples). By inspecting the prediction estimates shown in Supplementary Fig.~S2, a single combination of features (i.e. model order) potentially possessing the best parsimonious properties was then chosen for each feature set and each outcome (identified as circles in Supplementary Fig.~S2). The final logistic regression coefficients of these selected radiomic prediction models (3 feature sets $\times$ 3 outcomes) were then found by averaging all coefficients computed from another set of 100 bootstrap samples. These prediction models in their final form were thereafter directly tested in the defined testing set of this work (H\&N3 and H\&N4 cohorts; $n = 106$).

\subsection*{Combination of radiomic and clinical variables}
The construction of prediction models combining radiomic and clinical variables was also carried out using the training set consisting of the combination of 194 patients from the H\&N1 and H\&N2 cohorts (Fig.~\ref{fig:MLstrategy}). First, random forest classifiers \autocite{breiman_random_2001} containing only the following clinical variables were constructed for the LR, DM and OS outcomes: I) \textit{Age}; II) \textit{H}\&\textit{N type} (oropharynx, hypopharynx, nasopharynx or larynx); and III) Tumour stage. The selection of the following best groups of tumour stage variables to be incorporated into the \enquote{clinical-only} random forest classifiers was performed: I) \textit{T-Stage}; II) \textit{N-Stage}; III) \textit{T-Stage} and \textit{N-Stage}; and IV) \textit{TNM-Stage}. Estimation of prediction performance for feature selection and subsequent random forest training was performed in the training set using stratified random sub-sampling and imbalance adjustments to account for the disproportion between the occurrence and non-occurrence of events. Overall, the following staging variables were estimated to possess the highest prediction performance in the training set when combined into random forest classifiers with \textit{Age} and \textit{H}\&\textit{N type}, and were thereafter used for the rest of the work accordingly for each outcome: I) \textit{T-Stage} and \textit{N-Stage} for LR prediction; II) \textit{N-Stage} for DM prediction; and III) \textit{T-Stage} and \textit{N-Stage} for OS prediction. Finally, the variables of the previously identified radiomic prediction models (3 feature sets $\times$ 3 outcomes) were incorporated with the corresponding clinical variables identified for each outcome via the separate construction of final random forests classifiers.

\subsection*{Imbalance-adjustment strategy}
To obtain models with predictive power equally balanced between the prediction of occurrence of events and non-occurrence of events, an imbalance adjustment strategy adapted from the work of Schiller \textit{et al.} \autocite{schiller_modeling_2010} was used in this work (Supplementary Fig.~S4). Imbalance-adjustments become an essential part of the training process when the proportion of instances (e.g. patients) of a given class (e.g. occurrence of an event) is much lower than the proportion of instances of the other class (e.g.  non-occurrence of an event). This is the case in this work for the proportion of locoregional recurrences, distant metastases and death events in the training and testing sets (Fig.~\ref{fig:MLstrategy}).

In this work, every time a different bootstrap sample was drawn from the training set to construct a logistic regression or a random forest classifier, a different ensemble of multiple balanced classifiers was used in the training process instead of using only one unbalanced classifier. The ensemble classifier is composed of a number of $P = [N\mathrm{-}/N\mathrm{+}]$ partitions, where $N-$ is the number of instances from the majority class and $N+$ the number of instances from the minority class in a particular bootstrap sample. The $N+$ instances are copied and used in every partition, and the $N-$ instances are randomly sampled without replacement in the $P$ partitions such that the number of instances of the majority class is either $\lfloor N\mathrm{-}/P \rfloor$ or $\lceil N\mathrm{-}/P \rceil$ in each partition. For example, for $N\mathrm{-} = 168$ and $N\mathrm{+} = 32$, five partitions would be created: two would contain 33 instances from the majority class, three would contain 34 instances from the majority class, and all would contain the 32 instances from the minority class.

For the logistic regression training process, a different classifier (i.e. different coefficients) is then trained for each of the created partitions, and the final ensemble classifier consists in the average of the corresponding coefficients from each partition. For random forest training, each partition is used to create a decision tree to be appended to a final forest instead of creating only one tree per bootstrap sample.

\subsection*{Random forest training}
The process of random forest training inherently uses bootstrapping in order to train the multiple decision trees of the forest. Conventionally, one different decision tree is trained for each bootstrap sample. In this work, we used 100 bootstrap samples to train each random forest constructed from the training set (H\&N1 and H\&N2 cohorts; $n = 194$). For each bootstrap sample, the imbalance-adjustment strategy detailed above was used such that each bootstrap sample produced multiple decision trees (one per partition) to be appended to a random forest. Therefore, the final number of decision trees per random forest was dependent on the actual proportion of events in each bootstrap sample for each outcome studied. The three final random forest models developed in this work (italic fonts in Table~\ref{tab:comparison}, Supplementary Table~S4) were constructed using 582, 661 and 518 decision trees for LR, DM and OS, respectively.

In addition to the imbalance-adjustment strategy adopted in this work, under/oversampling of the instances in each partition of an ensemble was used to further correct for data imbalance in the random forest training process. Under/oversampling weights of the minority class of 0.5 to 2 with increments of 0.1 were tested in this work. Stratified random sub-sampling was used to estimate the optimal weight for a given training process (and also to estimate the optimal clinical staging variables to be used) in terms of the maximal average AUC, a process randomly separating the training set of this work into multiple sub-training and sub-testing sets ($n = 10$) with 2:1 size ratio and equal proportion of events. The final random forest models developed in this work (italic fonts in Table~\ref{tab:comparison}, Supplementary Table~S4) used oversampling weights of 1.4, 1.6 and 1.7 (in conjunction with the previously described imbalance-adjustment strategy) to train the decision trees of the forests for LR, DM and OS, respectively. The overall random forest training process is pictured in Supplementary Fig.~S7.

\subsection*{Calculation of prediction/prognostic performance metrics}
In this work, all prediction models were fully trained in the defined training set of this work (H\&N1 and H\&N2 cohorts; $n = 194$). Models were then independently tested in the defined testing set of this work (H\&N3 and H\&N4 cohorts; $n = 106$). Prediction performance was assessed using ROC metrics in terms of the AUC, sensitivity, specificity and accuracy of classification of binary clinical endpoints (locoregional recurrences, distant metastases, deaths). Prognostic performance in terms of time estimates of clinical endpoints was assessed using the concordance-index (CI) \autocite{harrell_multivariable_1996} and the \textit{p}-value obtained from Kaplan-Meier analysis using the log-rank test between risk groups.

For prediction performance, the output of the linear combination of features of logistic regression models was directly used to calculate the AUC with binary outcome data. The multivariable response was then transformed into the posterior probability of occurrence of an event using a logit transform to calculate the sensitivity, specificity and accuracy of prediction using a probability threshold of 0.5. Similarly, the output probability of occurrence of an event of random forest models was directly used to calculate the AUC with binary outcome data, and an output probability of 0.5 was also used to calculate the remaining metrics.

For prognostic performance, the output of the linear combination of features of cox proportional hazard regression models was directly used to calculate the CI with time-to-event data (time elapsed between the date the treatment ended and the date when an event occurred or the date of last-follow-up). The median of the output of the cox regression models found in the training set was used to separate the patients of the testing set into two risk groups for Kaplan-Meier analysis. For random forests, the output probability of occurrence of an event was directly used to calculate the CI with time-to-event data, and a probability threshold of 0.5 was used to separate the patients of the testing set into two risk groups (or $\frac{1}{3}$ and $\frac{2}{3}$ for three risk groups) for Kaplan-Meier analysis.

\subsection*{Code and models availability}
All software code used to produce the results presented in this work is freely shared under the GNU General Public License on the GitHub website at: \\ \url{https://github.com/mvallieres/radiomics}. 

Notably, a single organized script allowing to run all the experiments performed in this work is available, as well as a standalone MATLAB$^{\mathrm{\textregistered}}$ application in the form of a graphical user interface (GUI) to test the final models identified in this work on new patients.
\newpage

\begin{singlespacing}
\printbibliography
\end{singlespacing}
\newpage

\section*{Acknowledgements}
Martin Valli\`eres acknowledges partial support from the Natural Sciences and Engineering Research Council (NSERC) of Canada (scholarship CGSD3-426742-2012), and from the CREATE Medical Physics Research Training Network of NSERC (Grant number: 432290). We would also like to thank Fran\c cois Deblois from HGJ, Vincent Hubert-Tremblay, Luc Ouellet and Isabelle Gauthier from CHUS, and Jean-Fran\c cois Carrier, Chantal Boudreau and Karim Zerouali from CHUM for their help in retrieving the data at their respective institutions. Special thanks to Aditya Kumar for his help in the initiation of the project.

\section*{Author contributions}
M.V., J.S. and I.E.N. conceived the project, analyzed the data and wrote the paper. E.K-R., L.J.P, X.L., C.F., N.K., F.N., C.-S.W. and K.S. collected and curated the data. H.J.W.L.A. provided expert knowledge and helped in the computation of the radiomic signature. All authors edited the manuscript.

\section*{Competing financial interests}
The authors declare no competing financial interests.
\newpage

\end{document}